\documentclass[letterpaper, 10 pt, journal, twoside]{ieeetran}

\usepackage{graphicx}
\usepackage{amsmath,amsfonts}
\usepackage{array}
\usepackage{textcomp}
\usepackage{stfloats}
\usepackage{url}
\usepackage{verbatim}
\usepackage{graphicx}
\def\BibTeX{{\rm B\kern-.05em{\sc i\kern-.025em b}\kern-.08em
    T\kern-.1667em\lower.7ex\hbox{E}\kern-.125emX}}
\usepackage{balance}

\usepackage{algorithm}
\usepackage{algpseudocode}
\algrenewcommand\algorithmicrequire{\textbf{Input:}}
\algrenewcommand\algorithmicensure{\textbf{Output:}}

\usepackage{cite}
\usepackage{tikz}
\usepackage{pgfplots}
\usepackage{xcolor}
\pgfplotsset{compat=1.18}
\usetikzlibrary{positioning}
\usetikzlibrary{decorations.pathreplacing}
\usetikzlibrary{calc}
\usetikzlibrary{positioning,shapes,shadows,arrows}

\usepackage{pifont}
\usepackage{multirow}
\usepackage{dsfont}
\usepackage{subcaption}
\usepackage{bm}
\usepackage[font=footnotesize]{caption}

\usepackage[switch]{lineno}

\pgfplotsset{
every axis/.append style={
  axis line style={->}, 
  legend style={font=\scriptsize},
  label style={font=\scriptsize},
  title style={font=\scriptsize},
  tick label style={font=\scriptsize},
  }
}

\makeatletter
\let\NAT@parse\undefined
\makeatother
\usepackage{hyperref}

\begin{document}

\title{Collision-inclusive Manipulation Planning for Occluded Object Grasping via Compliant Robot Motions}

\author{Kejia Ren$^1$, 
        Gaotian Wang$^1$, 
        Andrew S. Morgan$^2$, 
        and Kaiyu Hang$^1$%
\thanks{This project is supported by the U.S. National Science Foundation under grant FRR-2240040.}
\thanks{$^{1}$Kejia Ren, Gaotian Wang, and Kaiyu Hang are with the Department of Computer Science, Rice University, Houston, TX 77005, USA (e-mail: kr43@rice.edu; gw23@rice.edu; kaiyu.hang@rice.edu)}
\thanks{$^{2}$Andrew S. Morgan is with RAI Institute, Cambridge, MA 02142, USA (e-mail: andy@rai-inst.com)}
}

\maketitle

\begin{abstract}
Robotic manipulation research has investigated contact-rich problems and strategies that require robots to intentionally collide with their environment,
to accomplish tasks that cannot be handled by traditional collision-free solutions.
By enabling compliant robot motions, collisions between the robot and its environment become more tolerable and can thus be exploited, but more physical uncertainties are introduced.
To address contact-rich problems such as occluded object grasping while handling the involved uncertainties, 
we propose a collision-inclusive planning framework that can transition the robot to a desired task configuration via roughly modeled collisions absorbed by Cartesian impedance control.
By strategically exploiting the environmental constraints and exploring inside a manipulation funnel formed by task repetitions,
our framework can effectively reduce physical and perception uncertainties.
With real-world evaluations on both single-arm and dual-arm setups, we show that our framework is able to efficiently address various realistic occluded grasping problems where a feasible grasp does not initially exist.
\end{abstract}


\section{Introduction}
\label{sec:intro}

\begin{figure}[t]
    \vspace{-3pt}
    \centering
    \begin{tikzpicture}
        \node[anchor=south west,inner sep=0] at (0,0){\includegraphics[width=\columnwidth]{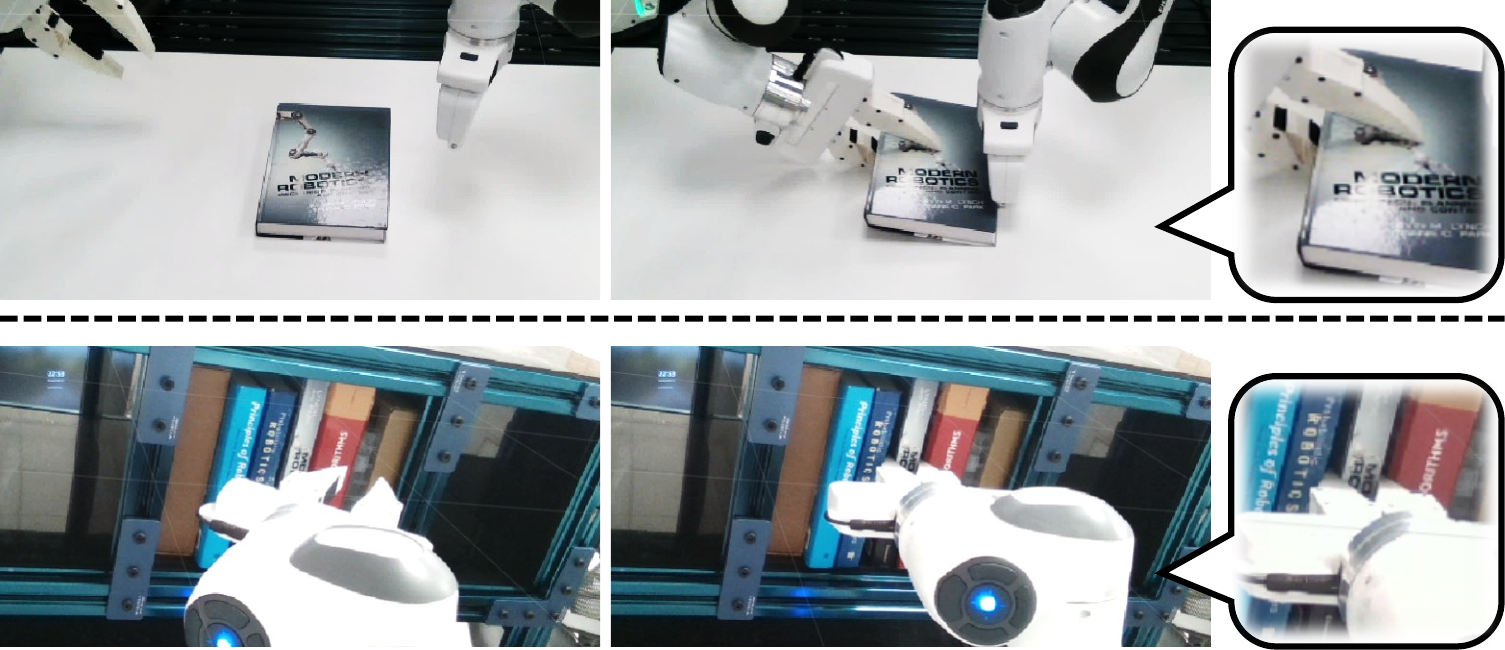}};
        \node[font=\footnotesize, anchor=west, align=left] at (7.25, 3.8) {Final Grasp};
        \node[font=\footnotesize, anchor=west, align=left] at (7.25, 1.761) {Final Grasp};
    \end{tikzpicture}
    \caption{Illustrations of occluded grasping tasks solved by our collision-inclusive planner.
    \emph{Top:} A book is lying flat on a table surface, needing the robot gripper to scoop-grasp it by contacting the table.
    \emph{Bottom:} A target book is tightly surrounded by other books on the shelf, requiring the robot gripper to insert its fingers into the gaps between books for a stable grasp.
    }
    \label{fig:first}
\end{figure}

Different from traditional collision-free solutions where contacts between robots and the environment are prohibited~\cite{wang2019manipulation, xiang2024grasping},
recent research in robotic manipulation has investigated how robots can make effective collisions with the environment to reach desired task configurations in various contact-rich problems~\cite{suomalainen2022survey}.
For a grasping example shown in Fig.~\ref{fig:first},
due to occlusions by surrounding objects,
a feasible grasp configuration for the target object does not initially exist,
and thus the robot has to collide with surrounding objects for a successful grasp.
Like how humans grasp in such configurations, by allowing collisions and safe interactions between the gripper and the environment,
the robot can insert its fingers into narrow gaps between objects to create free space for a desired grasp.

Such collision-inclusive strategies require safe contact and effective interactions between the robot and the objects,
for which robot compliance is essential~\cite{eppner2015exploitation}.
More recently, by novel hand designs with soft material and actuators~\cite{odhner2014compliant, deimel2016novel} or impedance-controlled motion strategies~\cite{balatti2018self}, robot compliance has been more extensively investigated and exploited for contact-rich manipulation.
Although compliance enables robots to have more manipulation possibilities, it also introduces more uncertainties to make the task execution more challenging.
In fact, exploiting environmental constraints (e.g., using contacts via compliance) can compensate for uncertainties to generate more robust manipulation outcomes~\cite{eppner2015planning}.
For the example scenarios shown in Fig.~\ref{fig:first},
with inaccurate object pose estimation and uncertain robot execution of compliant motions, directly placing the fingertip at the narrow gap and precisely navigating it through the gap to reach a desired grasp pose is almost impossible.
Instead, by maintaining continuous contact with the environment (e.g., the table surface or the surrounding books), the motion of the robot gripper will be constrained by the environment, and the transition of system state will become more certain;
as a result, the finger will successfully slide into the target gap while moving along the environmental surfaces.

In addition, the idea of \emph{Manipulation Funnels} has been studied to robustly manipulate objects against uncertainties, e.g., the variations in initial task configurations~\cite{mason1985mechanics}.
In this work, inspired by a broadened interpretation of manipulation funnels, we leverage repeated actions to produce stable manipulation outcomes under uncertainties.
With minor variations across different trials, task repetitions conceptually explore in a manipulation funnel to effectively guide the manipulation through desired motion transitions and find the solution.
More details will be discussed in Sec.~\ref{sec:prob}.

Motivated by the insights above, 
we propose a framework for occluded object grasping,
for which all feasible grasps are initially blocked by the environment.
Given mesh models of the robot gripper and objects, our framework is able to:
\begin{enumerate}
    \item 
    move the robot to its desired task configuration through trajectory generated by collision-inclusive planning;

    \item 
    generate effective compliant motions that exploit environmental constraints to reduce task uncertainties against inaccurate system modeling and perception;

    \item 
    produce more stable task outcomes through exploring a manipulation funnel formed by repeated actions.
\end{enumerate}

\section{Related Work}

\emph{Contact-rich Manipulation.}
Robotic manipulation that requires contact with the environment has been extensively studied for different tasks and applications,
such as wiping or polishing~\cite{amanhoud2020force}, peg-in-hole~\cite{guan2018efficient, nottensteiner2020robust}, assembly~\cite{almeida2016folding}, etc.
Based on different task requirements and sensing capabilities, the strategies for planning and controlling robot motions need to incorporate different functionalities, such as estimation of contact state, control of contact forces, tolerance to inaccurate modeling, etc.
Our framework uses a highly approximate system model to plan compliant motions, where controlling contact force is unnecessary for the tasks of our interest.

\emph{Occluded Object Grasping.}
When all feasible grasp configurations are occluded and
traditional collision-free frameworks cannot solve the problem,
solutions that actively leverage contacts to reconfigure objects are needed.
Nonprehensile pre-grasp strategies such as pushing-based rearrangement~\cite{dogar2011framework, lee2019efficient}, sliding~\cite{king2013pregrasp, hang2019pre, liang2021learning} and flipping~\cite{sun2020learning, zhou2023learning, yang2024learning} are developed via manually designed primitives or learned policies.
In general,
planning-based approaches~\cite{cheng2023enhancing} are more generalizable but rely on explicit contact modeling, making them sensitive to real-world uncertainties; 
while learning-based methods~\cite{chen2023synthesizing, zhang2023reinforcement} are more robust to such uncertainties but typically require large amounts of data to generalize.
Unlike these existing methods which require significant reconfiguration of objects, our framework is effective in confined spaces without needing much object motions.

\emph{Exploitation of Environmental Constraints.}
Contacts with the environment can be leveraged as environmental constraints to help mitigate uncertainties~\cite{sieverling2017interleaving}.
This concept has been successfully applied in various manipulation problems to generate robust solutions in grasping~\cite{eppner2015planning, pall2021analysis} and in-hand manipulation~\cite{dafle2014extrinsic}.
Furthermore, environmental constraints can also facilitate the learning of manipulation skills~\cite{subramani2018inferring, shao2020learning, li2022learning} for improved robustness and generalizability.
In this work,
we exploit environmental constraints in the form of a trajectory of contacts,
to reduce the uncertainties caused by inaccurate sensing and approximate dynamics modeling.

\section{Problem Statement}
\label{sec:prob}

In this work, we are interested in planning compliant motions of a robot manipulator in a collision-rich environment, 
to grasp a target object whose stable grasping configuration is originally occluded by other objects in the environment.
For such reasons, the robot has to 1) physically interact with the environment by making safe collisions with it;
2) create free space without greatly changing the environment configuration;
and 3) insert its gripper into the desired grasp pose which is originally infeasible due to occlusions.

\subsection{Preliminaries}
\label{sec:term}

The configuration of a robot at time $t$ is represented by its joint angles $\bm{q}_t \in \mathbb{R}^N$ where $N$ is the degrees-of-freedom (DoF) of the robot.
We specify a workspace with respect to the robot's base frame.
A task frame is defined on the robot gripper and we denote the $SE(3)$ pose of this frame in the workspace (i.e., the robot's base frame) by $\bm{x}_t \in \mathbb{R}^6$ (i.e., position and orientation represented by roll, pitch, and yaw angles), which can be obtained by forward kinematics of the robot $\bm{x}_t = \mathrm{FK}(\bm{q}_t)$.
Furthermore, We use the capital notation $\bm{X}_t \in SE(3)$ to represent the pose of the task frame (i.e., $\bm{x}_t$) as a transformation matrix.
Given a point $\bm{p} \in \mathbb{R}^3$ in the task frame, the operation $\bm{X}_t \cdot \bm{p}$ calculates transformed coordinates of this point in the robot's base frame.
In this work, since the occluded grasping tasks require the robot to insert its fingers into narrow gaps between objects, the task frame is defined at one of the fingertips of the gripper.

An impedance control (commanded to the robot) is modeled by a reference equilibrium pose $\bm{\xi} \in SE(3)$ for the task frame, and its associated stiffness and damping parameters $\bm{K}, \bm{D} \in \mathbb{R}^{6 \times 6}$.
By setting the equilibrium pose to $\bm{\xi}$ in the workspace, a Cartesian impedance controller virtually exerts a wrench $\bm{f}_t \in \mathbb{R}^6$ (including 3D force and torque) at the task frame (i.e., gripper's fingertip) to perform a robot Cartesian behavior mimicking a virtual spring-damping system:
\begin{equation}
    \bm{f}_t = \bm{K} \left( \bm{\xi} - \bm{x}_t \right) - \bm{D} \dot{\bm{x}}_t 
\end{equation}
where $\dot{\bm{x}}_t \in \mathbb{R}^6$ is the velocity of the task frame.
The robot state will then evolve under the impedance-controlled dynamics as follows:
\begin{equation}
\label{eq:dynamics_q}
\begin{aligned}
    \bm{M}\left( \bm{q}_t \right) \ddot{\bm{q}}_t + \bm{C}\left(\bm{q}_t, \dot{\bm{q}}_t \right) &= \bm{J}\left(\bm{q}_t \right) ^ \top \left( \bm{f}_t + \bm{f}_{ext} \right)
\end{aligned}
\end{equation}
where $\dot{\bm{q}}_t, \ddot{\bm{q}}_t \in \mathbb{R}^N$ are the joint velocities and accelerations of the robot; 
$\bm{M}\left(\bm{q}_t\right) \in \mathbb{R}^{N \times N}$ is the mass matrix; 
$\bm{C}\left(\bm{q}_t, \dot{\bm{q}}_t \right) \in \mathbb{R}^N$ contains the centrifugal and Coriolis terms;
$\bm{J}\left( \bm{q}_t \right) \in \mathbb{R}^{6 \times N}$ is the Jacobian matrix;
$\bm{f}_{ext} \in \mathbb{R}^6$ is an external wrench received by the robot at the task frame.

The task environment contains $M$ objects of interest $\left\{\mathcal{O}_1, \cdots, \mathcal{O}_M \right\}$, 
where $\mathcal{O}_i \subset \mathbb{R}^3$ represents the geometry of the $i$-th object.
The pose of the $i$-th object in the workspace (i.e., the robot's base frame) is denoted by $\bm{X}^i \in SE(3)$ as a transformation matrix.
One of the objects, $g \in \left\{1, \cdots, M \right\}$ is the target object to be grasped;
the geometry and pose of the target object are denoted by $\mathcal{O}^g \subset \mathbb{R}^3$ and $\bm{X}^g \in SE(3)$ respectively.
Given the geometry of the target object, a stable grasp pose $\bm{X}^* \in SE(3)$ for the task frame (of the robot gripper) can be obtained by an off-the-shelf grasp planning algorithm, which is expressed in the target object's body frame.
To express this grasp pose in the robot's base frame, we need to transform it by $\bm{X}^g \cdot \bm{X}^*$,
where the operator $\cdot$ is matrix multiplication for transformation matrices.

We aim to find a sequence of Cartesian impedance controls, defined by reference equilibrium poses $\left\{\bm{\xi}_0, \cdots, \bm{\xi}_{T-1}\right\}$ of task frame, to move the robot gripper to reach the desired (yet initially occluded) grasp pose $\bm{X}^g \cdot \bm{X}^*$ from a starting collision-free pose $\bm{X}_0 \in SE(3)$ through safe contacts and interactions with the environment.
Unlike methods (e.g., ~\cite{cheng2023enhancing, chen2023synthesizing, zhang2023reinforcement}) that actively change object poses, our framework makes minimal changes to object configurations (i.e., $\bm{X}_i$),
which is particularly effective when object motions are highly constrained (e.g., in a confined space with tightly packed objects,
or when object motions can be blocked by another robot arm or environmental constraints).

\begin{figure}[t]
    \centering
    \includegraphics[width=\columnwidth]{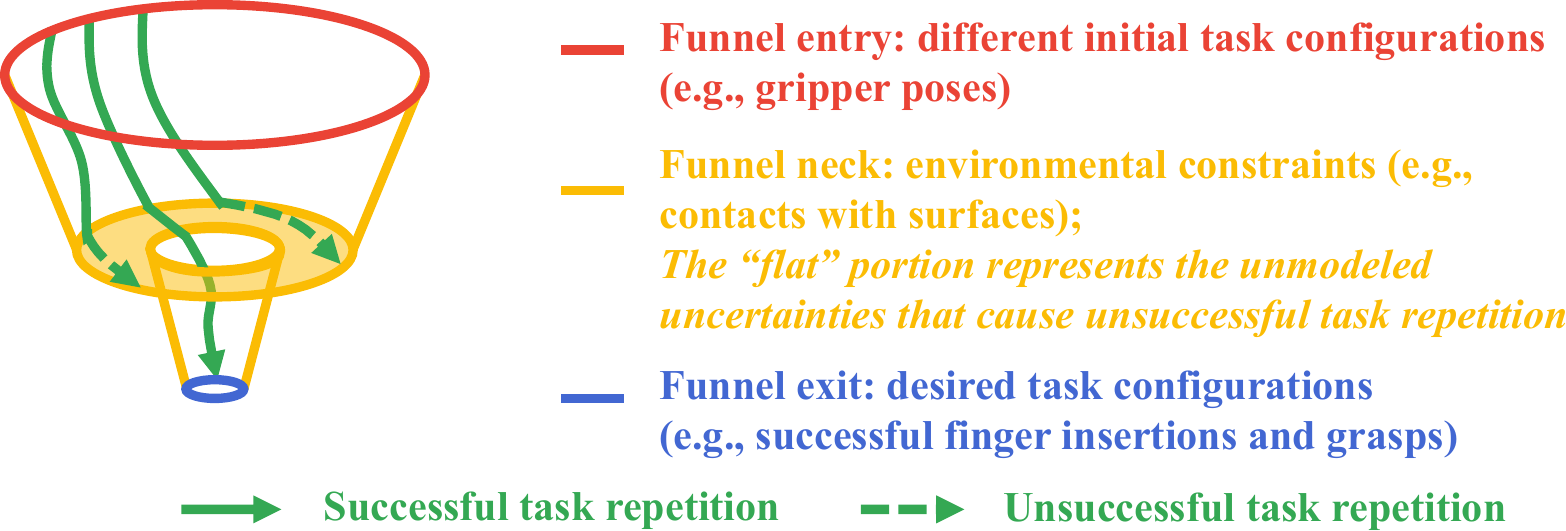}
    \vspace{-15pt}
    \caption{Degenerate manipulation funnel by repetitions.
    }
    \label{fig:rep_funnel}
    \vspace{-15pt}
\end{figure}

\begin{figure}[t]
\begin{algorithm}[H]
\small
\caption{Collision-inclusive Manipulation Planning}
    \begin{algorithmic}[1]
        \Require Object geometries $\left\{ \mathcal{O}_1, \cdots, \mathcal{O}_M \right\}$ where $g \in \{1, \cdots, M\}$ is the index of the target object, desired grasp pose of the robot gripper's task frame $\bm{X}^*$ specified in the target object's body frame, maximum number of repetitions $K_{\max}$
        \Ensure Task success ($true$ or $false$)
        \State $k \gets 0$ \Comment{Number of Repetitions}
        \While{$k < K_{\max}$}
            \State $k \gets k + 1$
            \State $\left \{\bm{X}^1, \cdots, \bm{X}^M \right\} \gets \Call{Observe}{\null}$ \Comment{Object Poses}
            \State $\bm{X}_0 \gets \Call{SampleAround}{\bm{X}^g \cdot \bm{X}^*}$
            \State $\left\{\widetilde{\bm{X}}_t\right\}_{t=1}^T \gets \Call{GeomPath}{\bm{X}_0, \bm{X}^g \cdot \bm{X}^*}$ \Comment{Sec.~\ref{sec:geom_path}}
            \State $\left\{\bm{\xi}_t\right\}_{t=0}^{T-1} \gets \Call{ImpdCtrl}{\left\{\widetilde{\bm{X}}_t \right\}_{t=1}^T}$            \hfill \Comment{Sec.~\ref{sec:impd_ctrl}}
            \State $\Call{MoveGripperTo}{\bm{X}_0}$
            \If{$\Call{Execute}{\left\{\bm{\xi}_0, \cdots, \bm{\xi}_{T-1} \right\}}$}
                \State \Return $true$
            \EndIf
        \EndWhile
        \State \Return $false$
    \end{algorithmic}
\label{alg:repetition}
\end{algorithm}
\vspace{-20pt}
\end{figure}

\vspace{-5pt}
\subsection{Exploring in Manipulation Funnel by Repetitions}

The impedance controls are generated in an open loop and may not lead to successful task completion during real executions, due to physical and perception uncertainties.
However, 
as failed execution does not greatly change the environment configuration,
a repeated execution with some local randomness is still likely to achieve the task. 
To this end, our framework allows the robot to attempt more trials with repeated actions.
Conceptually, as shown in Fig.~\ref{fig:rep_funnel}, 
we can formulate a degenerate manipulation funnel that allows repeated actions to enter and navigate through.
The entry of the funnel is defined by all initial task configurations (e.g., gripper poses, minor variations in object shape and poses);
the exit of the funnel is all desired task completion states (e.g., stable grasps);
the neck is the environmental or other task-implicit constraints that transition the manipulation towards the desired states.
Each task repetition planned using environmental constraints is a series of motions that go inside the funnel from its entry towards the exit.
Due to inaccurate modeling and uncertainties (as represented by the ``flat'' degenerate funnel neck in the figure),
a repetition may not successfully find the funnel exit.
However, the repetitions are always contained inside the funnel by not greatly disrupting the environment configuration and constraints.
With more valid repetitions, 
the manipulation will finally locate the funnel exit to robustly finish the task.

Our proposed framework is outlined in Alg.~\ref{alg:repetition}.
Specifically, 
the framework first observes the poses of objects in the environment.
Provided with a desired grasp pose $\bm{X}^*$, a starting collision-free pre-grasp pose $\bm{X}_0$ of the gripper will be sampled around the workspace.
To move the task frame of the robot gripper from $\bm{X}_0$ to the desired grasp $\bm{X}^g \cdot \bm{X}^*$,
our collision-inclusive planner will then generate a sequence of desired impedance controls in two steps:
1) by exploiting environmental constraints (e.g., contacts with the objects or support surfaces), a geometric path will be planned for the task frame in Sec.~\ref{sec:geom_path};
2) Guided by the geometric path planned in the previous step, a sequence of impedance controls, represented by equilibrium poses for the task frame, will be generated via optimization in Sec.~\ref{sec:impd_ctrl}.
The generated impedance controls will be commanded to the robot for execution.
Meanwhile, if the task frame greatly deviates from the planned geometric path in Step 1), we will early terminate the execution by returning a $false$ in the $\Call{Execute}{\cdot}$ function and skip to the next trial.
After all impedance controls have been executed, the robot will close its gripper to grasp the target object.
If the grasp is stable, as assessed by some off-the-shelf grasping criteria, the task is completed;
otherwise, the robot will open the gripper, move to a newly sampled pre-grasp pose $\bm{X}_0$, and start the next trial.
The repetition continues until the target object is stably grasped or the number of trials exceeds a preset limit $K_{\max}$.
\section{Collision-inclusive Planning}
\label{sec:method}

Given the geometric modeling of the robot gripper and environmental objects as will be described in Sec.~\ref{sec:geometric_repre}, 
this section introduces the procedure and details how the framework generates impedance controls through optimization-based planning, as will be detailed in Sec.~\ref{sec:geom_path} and~\ref{sec:impd_ctrl}.

\begin{figure}[t]
    \centering
    \begin{tikzpicture}
        \node[anchor=south west,inner sep=0] at (0,0){\includegraphics[width=0.9\columnwidth]{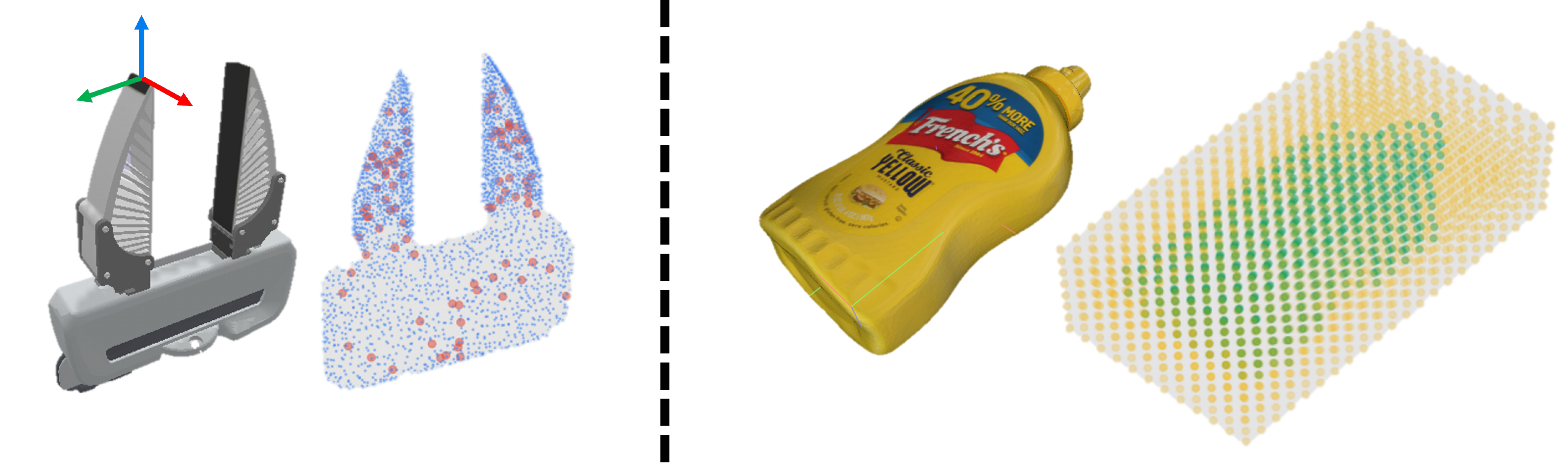}};
        \node[font=\footnotesize, anchor=west, align=left] at (0.75, 2.25) {Task Frame};
    \end{tikzpicture}
    \vspace{-5pt}
    \caption{
    \emph{Left:}
    The gripper's geometry is represented by a point cloud:
    $1000$ points sampled on the surface (blue) and $50$ inside the volume (red) of the mesh model.
    The task frame is defined at one fingertip.
    \emph{Right:} The object (i.e., mustard bottle) is modeled by a signed distance field (SDF), where positive and negative values are visualized by yellow and green points respectively.
    }
    \label{fig:task_frame}
    \vspace{-15pt}
\end{figure}

\subsection{Geometric Representations of Robot and Objects}
\label{sec:geometric_repre}

Approximate geometric representations are needed for efficient and differentiable distance computation between the robot and environment.
As shown in Fig.~\ref{fig:task_frame} (left),
we approximately represent the geometry of the robot gripper by a point cloud sampled on the gripper's mesh model, i.e., $\mathcal{P} \subset \mathbb{R}^3$, where the coordinates of each point $\bm{p} \in \mathcal{P}$ is given in the task frame on the gripper.
The points are sampled both on the surface and inside the volume of the gripper's mesh model, i.e., $\mathcal{P} = \mathcal{P}^s \cup \mathcal{P}^v$.
The surface points $\mathcal{P}^s$ should be sufficiently dense to capture the robot's contact geometry, 
while the internal points $\mathcal{P}^v$ are kept sparse to reduce the computational cost of collision optimization.

For each object $\mathcal{O}_i$, we generate a discrete signed distance field (SDF).
We first sample a 3D grid with a resolution of $\Delta$ around the object;
then, we compute the signed distance value for each voxel of the grid to construct a discrete SDF, as shown in Fig.~\ref{fig:task_frame} (right).
Given an arbitrary query point $\bm{p} \in \mathbb{R}^3$ in the object's body frame, 
the signed distance of this query point can be approximated by interpolating values on the discrete SDF grid, which we denote by $\mathrm{SDF}_i (\bm{p}) \in \mathbb{R}$.
To infer the signed distance of a point $\bm{p}$ given in the robot's base frame,
we define a function $\phi_i: \mathbb{R}^3 \mapsto \mathbb{R}$, which queries the SDF after transforming this point into the object's body frame using the observed object's pose $\bm{X}^i \in SE(3)$:
\begin{equation}
    \phi_i(\bm{p}) = \mathrm{SDF}_i \left( \left( \bm{X}^i \right)^{-1} \cdot \bm{p} \right)
\end{equation}
In addition, 
to obtain the signed distance of a point $\bm{p}$ (given in the robot's base frame) to the whole environment consisting of all objects, 
we define another function $\phi: \mathbb{R}^3 \mapsto \mathbb{R}$ by taking the signed distance to its nearest object:
\begin{equation}
    \phi(\bm{p}) = \min_{i \in \left\{1, \cdots, M\right\}} \phi_i(\bm{p})
\end{equation}

\subsection{Constraint-Exploiting Geometric Path Planning}
\label{sec:geom_path}

\begin{figure}[t]
    \centering
    \begin{tikzpicture}
        \node[anchor=south west,inner sep=0] at (0, 0){ \includegraphics[width=0.93\columnwidth]{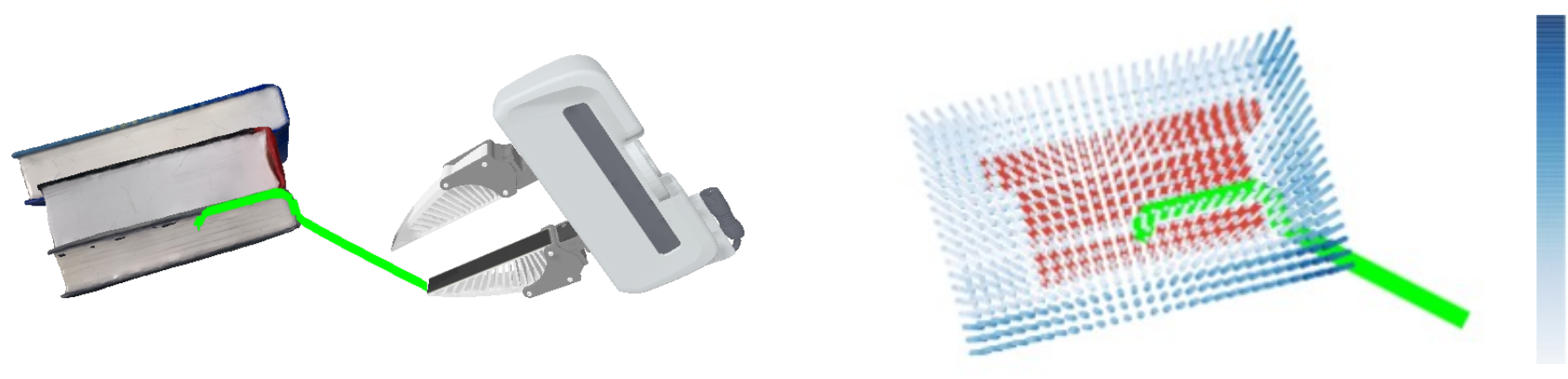}};
        \node[font=\footnotesize, anchor=west, align=left] at (8.2, 1.76) {9};
        \node[font=\footnotesize, anchor=west, align=left] at (8.2, 1.24) {4};
        \node[font=\footnotesize, anchor=west, align=left] at (8.2, 0.72) {1};
        \node[font=\footnotesize, anchor=west, align=left] at (8.2, 0.2) {0};
    \end{tikzpicture}
    \vspace{-5pt}
    \caption{Positional path-finding via $A^*$ algorithm illustrated with a book-grasping example.
    \emph{Left:} 
    The fingertip of the gripper (i.e., the task frame) needs to follow a desired positional path (green),
    to exploit environmental constraints (i.e., contacts with object surfaces) for a more robust insertion into the gap between books.
    \emph{Right:} 
    The desired positional path (green) is planned on the 3D discretized workspace using an $A^*$ algorithm.
    The red voxels penetrate the environment by a preset threshold $\delta_g$ and are regarded as obstacles.
    The colors of the blue voxels (i.e., non-obstacle voxels) represent their squared signed distance values, i.e., $\phi\left(\bm{p}\right)^2$, in $cm^2$.
    }
    \label{fig:geom_path}
\vspace{-14pt}
\end{figure}

Precisely placing the robot finger to a narrow gap for pre-grasp insertion merely based on visual perception is not reliable due to perceptual uncertainty.
Instead, locally exploring and exploiting the environmental constraints (e.g., touching and sliding on the environmental surfaces) is more robust to such uncertainties for locating the gaps.
For example, when inserting a finger into the gap between two tightly stacked books as shown in Fig.~\ref{fig:geom_path},
directly hitting the finger toward the gap from in the air based on purely visual perception is likely to miss it since the gap's location can be inaccurately observed.
However, 
if we let the robot slide its finger on the book surface near the gap, 
the robot can more robustly find the gap by trapping its finger in the gap during sliding.
Based on this intuition, we expect the task frame (at the fingertip) to follow a geometric path that actively exploits environmental constraints by contacting object surfaces, 
as shown in Fig.~\ref{fig:geom_path}.
Meanwhile, 
we also want to reduce collisions with other parts (e.g., the gripper base) of the robot gripper, 
since unexpected contacts can easily deviate the gripper from its desired motion behavior while the robot moves compliantly.
Considering both aspects,
we propose to generate a desired geometric path for the gripper's task frame in two steps:

\subsubsection{Positional Path-Finding via $A^*$ Algorithm}
\label{sec:geom_astar}

As the first step, we plan only the positional motions of the fingertip, which will be used as a guidance to plan the entire pose path (i.e., position and orientation) in the next step.
To exploit environmental constraints, we want the task frame (fingertip) to get close enough to the environmental surface and contact it,
but not to penetrate too much into the environment to be performed in reality.
Particularly, the positional path-finding for the task frame is formulated as an optimization problem:
\begin{subequations}
\label{eq:opt1}
\begin{align}
    \underset{\bm{p}_1, \cdots, \bm{p}_T}{\textrm{minimize}} \quad & w \cdot \sum_{t=1}^T \phi\left( \bm{p}_t \right)^2 + \sum_{t=0}^{T-1} \lVert \bm{p}_t - \bm{p}_{t+1} \rVert^2 \label{eq:opt1_a}\\
    \textrm{subject to} \quad & \phi \left(\bm{p}_t \right) + \delta_g \leq 0, \quad \forall t = 1, \cdots, T \label{eq:opt1_b}
\end{align}
\end{subequations}
where the first term in the objective of Eq.~\eqref{eq:opt1_a} is squared SDF to encourage the fingertip to touch environmental surfaces, and the second term is a smoothing term;
$w$ is a scaling factor to trade-off between two terms in the objective, which we set to $w = 1 / \Delta^2$ for all our experiments ($\Delta$ is the grid resolution of SDF).
By setting a nonnegative threshold $\delta_g$ in the constraint Eq.~\eqref{eq:opt1_b}, we ensure the fingertip will not penetrate the environment too much.

We solve the above path-finding problem by searching via an $A^*$ algorithm over the 3D workspace, as illustrated in Fig.~\ref{fig:geom_path} (right).
Specifically, we discretize the workspace containing objects into a 3D grid with resolution $\Delta$.
For voxels whose coordinates penetrate the environment by a value greater than $\delta_g$, i.e., violating the constraint in Eq.~\eqref{eq:opt1_b}, we regard them as being an obstacle to prevent the path from passing through it.
For two neighbor voxels with coordinates $\bm{p}$ and $\bm{p}'$, the cost is defined as $c\left(\bm{p}, \bm{p}'\right) = w \cdot \phi\left (\bm{p}'\right)^2 + \lVert \bm{p} - \bm{p}' \rVert^2$; and the heuristic function is simply the smoothing term (i.e., the path length), i.e., $h\left(\bm{p}, \bm{p}'\right) = \lVert \bm{p} - \bm{p}' \rVert^2$, which underestimates the cost to ensure optimality of the found path.
If a valid path (a sequence of traversed grid voxels) is found,
we interpolate it with $T$ waypoints to extract a positional path $\left\{\bm{p}_1, \cdots, \bm{p}_T \right\}$ as the solution to Eq.~\eqref{eq:opt1}.

\begin{figure}[t]
\vspace{-10pt}
\begin{algorithm}[H]
\small
\caption{Iterative Path-Finding via $A^*$ Algorithm }
    \begin{algorithmic}[1]
        \Require Initial $\delta_g$, increment step $\,d\delta_g$
        \Ensure The found path $\left\{ \bm{p}_1, \cdots, \bm{p}_T \right\}$
        \While{$true$}
            \State $path \gets \Call{AStarAlg}{\delta_g}$ \Comment{Over a 3D Grid Workpsace}
            \If{$path$ is not $null$}
                \State $\left\{ \bm{p}_1, \cdots, \bm{p}_T \right\} \gets \Call{Interpolate}{path}$
                \State \Return $\left\{ \bm{p}_1, \cdots, \bm{p}_T \right\}$
            \EndIf
            \State $\delta_g \gets \delta_g + \,d\delta_g$
        \EndWhile
    \end{algorithmic}
\label{alg:a_star}
\end{algorithm}
\vspace{-15pt}
\end{figure}

It is worth noting that the threshold $\delta_g$ for allowed penetration is important for the existence of a solution.
As detailed in Alg.~\ref{alg:a_star}, we run our $A^*$ with an initial $\delta_g$.
If the path does not exist, we will relax the allowed penetration $\delta_g$ by iteratively increasing it by $\,d\delta_g$, until a path is found.

\subsubsection{Optimization-based Path Refinement via Collision Minimization}
\label{sec:geom_refine}

In the second step, we aim to minimize collisions between the gripper and the environment, while still keeping the fingertip following the constraint-exploiting path found in the previous step.
To this end, in Eq.~\eqref{eq:g_v}, given the signed distance $\phi(\bm{p}) \in \mathbb{R}$ of a point $\bm{p} \in \mathbb{R}^3$, we define a collision cost as in CHOMP~\cite{ratliff2009chomp}. 
The margin parameter $\varepsilon$, set relative to object's size, helps prevent the gripper from moving too close to obstacles; 
a larger value of $\varepsilon$ is used when more conservative collision avoidance is encouraged,
such as in the presence of significant perception errors.
\begin{equation}
    c\left(\phi\left(\bm{p}\right)\right) =
    \begin{cases}
        -\phi\left(\bm{p}\right) + \frac{1}{2} \varepsilon & \phi\left(\bm{p}\right) < 0 \\
        \frac{1}{2 \varepsilon} \left(\phi\left(\bm{p}\right) - \varepsilon \right)^2 & 0 \leq \phi\left(\bm{p}\right) < \varepsilon \\
        0 & \text{otherwise}
    \end{cases} \label{eq:g_v}
\end{equation}
Given the pose $\bm{X}_t \in SE(3)$ of the task frame, the overall collision cost is computed by summing for each object $i$ and each sampled point $\bm{p} \in \mathcal{P}^v$ within the gripper's volume:
\begin{equation}
    C\left( \bm{X}_t \right) = \sum_{i = 1}^M \sum_{\bm{p} \in \mathcal{P}^v} c\left( \phi_i\left( \bm{X}_t \cdot \bm{p}\right) \right)
\end{equation}
where $\bm{X}_t \cdot \bm{p}$ transform a point $\bm{p} \in \mathcal{P}^v$ defined in the task frame to the robot's base frame.

Then, guided by the positional path $\left\{ \bm{p}_1, \cdots, \bm{p}_T \right\}$ found in the previous step,
we refine to obtain the geometric path $\left\{ \widetilde{\bm{X}}_1, \cdots, \widetilde{\bm{X}}_T\right\}$ where $\widetilde{\bm{X}}_t \in SE(3)$ (i.e., containing both positions and orientations), by minimizing the collision cost in a constrained optimization problem in Eq.~\eqref{eq:opt2}:
\begin{subequations}
\label{eq:opt2}
\begin{align}
    \underset{\widetilde{\bm{X}}_1, \cdots, \widetilde{\bm{X}}_T}{\textrm{minimize}} \quad & \sum_{t=1}^T C\left( \widetilde{\bm{X}}_t \right) + \sum_{t=0}^{T-1} d\left(\widetilde{\bm{X}}_t, \widetilde{\bm{X}}_{t+1} \right) \label{eq:opt2_a}\\
    \textrm{subject to} \quad & \widetilde{\bm{X}}_T = \bm{X}^g \cdot \bm{X}^* \label{eq:opt2_b}\\
                  \quad & \lVert \widetilde{\bm{p}}_t - \bm{p}_t \rVert \leq \delta_p, \quad \forall t \in \{1, \cdots, T\} \label{eq:opt2_c}
\end{align}
\end{subequations}
where the first term of the objective Eq.~\eqref{eq:opt2_a} is the collision cost;
the second is a distance smoothing term; 
the constraint in Eq.~\eqref{eq:opt2_b} is to ensure the final pose of the solution reaches the desired grasp pose; 
the constraint in Eq.~\eqref{eq:opt2_c} prevents the solution geometric path from deviating from the constraint-exploiting positional path $\left\{ \bm{p}_t\right\}_{t=1}^T$ found in the first step (note: $\widetilde{\bm{p}}_t \in \mathbb{R}^3$ denotes the positional component of $\widetilde{\bm{X}}_t$).

\subsection{Impedance Control Generation}
\label{sec:impd_ctrl}

\begin{figure*}[t]
    \centering
    \includegraphics[width=0.85\linewidth]{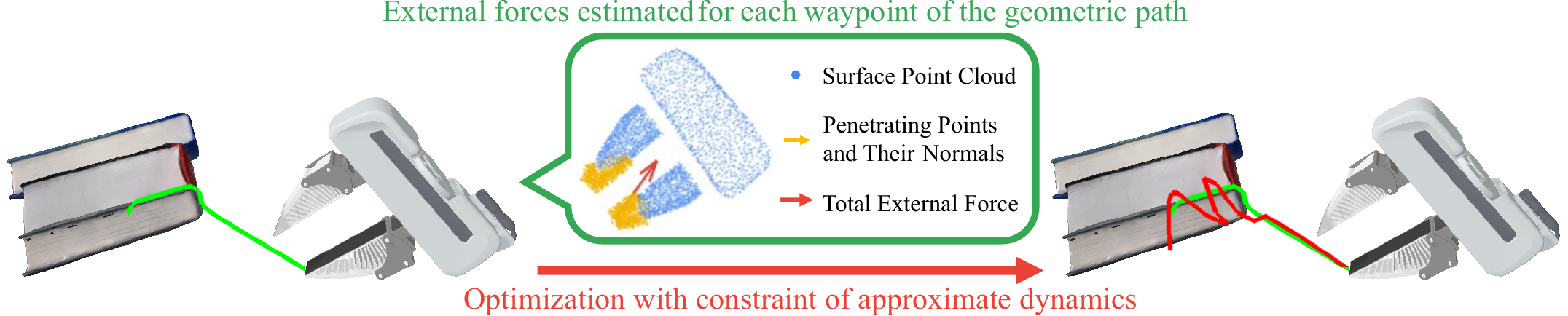}
    \caption{
    Illustration of impedance control generation with a book-grasping example.
    The left figure shows the geometric path (green lines) found by the approach described in Sec.~\ref{sec:geom_path}.
    In the middle figure, for each waypoint of the geometric path, an external force (red arrow) can be estimated by averaging the effects of penetrating points (yellow).
    The right figure shows the generated impedance controls (i.e., the red trajectory of equilibrium poses), solved by an optimization program constrained by the approximate dynamics.
    }
    \label{fig:impd_path}
    \vspace{-13pt}
\end{figure*}

After a desired geometric path has been planned in Sec.~\ref{sec:geom_path}, we need to find a sequence of impedance controls $\left\{ \bm{\xi}_0, \cdots, \bm{\xi}_{T-1} \right\}$ to drive the robot's gripper to follow the desired geometric path, by using robot dynamics as a heuristic.
In Eq.~\eqref{eq:dynamics_q}, the value of term $\bm{C}\left(\bm{q}_t, \dot{\bm{q}}_t \right)$ can be observed in real time; 
this term can be canceled from the dynamics equation by implementing an internal Coriolis-compensating mechanism for the controller.
Then, by substituting $\ddot{\bm{x}} = \bm{J} \ddot{\bm{q}} + \dot{\bm{J}} \dot{\bm{q}}$ into Eq.~\eqref{eq:dynamics_q} of robot dynamics,
the impedance-controlled behavior of the task frame can be inferred:
\begin{equation}
\begin{aligned}
    \bm{\Lambda}\left( \bm{q}_t \right) \left(\ddot{\bm{x}}_t - \dot{\bm{J}}\left( \bm{q}_t\right) \dot{\bm{q}}_t \right) &=  \bm{f}_t + \bm{f}_{ext}
\end{aligned}
\end{equation}
where $\bm{\Lambda}\left( \bm{q}_t \right) = \bm{J}\left(\bm{q}_t\right)^{-\top} \bm{M}\left(\bm{q}_t\right) \bm{J}\left(\bm{q}_t\right)^{-1}$ is the effective inertia matrix of robot in the task space.
When the robot moves slowly, the term $\dot{\bm{J}}\left( \bm{q}_t\right) \dot{\bm{q}}_t$ is neglected,
yielding a computationally efficient dynamics approximation:
\begin{equation}
    \ddot{\bm{x}}_t \approx \bm{\Lambda} \left( \bm{q}_t\right) ^ {-1} \left( \bm{f}_t + \bm{f}_{ext} \right)
\end{equation}
This approximation may be problematic for tasks involving dynamic, high-speed motions or rapid joint accelerations (e.g., near singularities).
However, as we always slowly move the robot in contact-rich actions,
our assumption of slow motion always holds in relevant tasks.

The external wrench $\bm{f}_{ext} \in \mathbb{R}^6$ is estimated based on the virtual penetration along the planned geometric path.
The estimate of $\bm{f}_{ext}$ is rough (and does not need to be accurate given other perception and modeling errors), yet remains informative for heuristically guiding the control generation.
Specifically, 
when the gripper is at pose $\widetilde{\bm{X}}_t$ while following the geometric path, 
a local force $\bm{f}\left( \bm{p} \right) \in \mathbb{R}^3$ can be estimated at each point $\bm{p} \in \mathcal{P}^s$ sampled on the gripper's surface.
The force $\bm{f}\left( \bm{p} \right)$ has a magnitude proportional to the penetration distance (i.e., the negative of signed distance) from this point to the environment;
the direction of $\bm{f}\left( \bm{p} \right)$ is the same as the normal direction of the gripper surface at this point:
\begin{equation}
    \bm{f}\left( \bm{p} \right) = \mathds{1}_{\left\{ \phi\left(\widetilde{\bm{X}}_t \cdot \bm{p}\right) < 0 \right\}}\cdot k \phi\left(\widetilde{\bm{X}}_t \cdot \bm{p}\right) \cdot \bm{n}
\end{equation}
where $\phi\left(\widetilde{\bm{X}}_t \cdot \bm{p}\right)$ is the signed distance from a point $\bm{p}$ (on the gripper's surface) to the environment;
the indicator function $\mathds{1}_{\left\{ \phi\left(\widetilde{\bm{X}}_t \cdot \bm{p}\right) < 0 \right\}}$ filters out the points not in penetration since such points will not exert forces;
$k$ is a hyperparameter corresponding to the stiffness of virtual penetrations;
$\bm{n} \in \mathbb{R}^3$ is the outward normal vector of the gripper's geometric model at point $\bm{p}$. 
Then, the estimated external wrench $\bm{f}_{ext}$ (i.e., force and torque) received by the gripper is obtained by averaging the effect of local force at each point:
\begin{equation}
    \bm{f}_{ext}\left(\widetilde{\bm{X}}_t \right) = \frac{1}{Z}
    \begin{pmatrix}
    \widetilde{\bm{R}}_t \cdot \sum_{\bm{p} \in \mathcal{P}^s} \bm{f}\left( \bm{p} \right)\\
    \widetilde{\bm{R}}_t \cdot \sum_{\bm{p} \in \mathcal{P}^s} \bm{p} \times \bm{f}\left( \bm{p} \right)
    \end{pmatrix} \in \mathbb{R}^6
\end{equation}
where $Z = \sum_{\bm{p} \in \mathcal{P}^s} \mathds{1}_{\left\{ \phi\left(\widetilde{\bm{X}}_t \cdot \bm{p}\right) < 0 \right\}}$ calculates the number of points in penetration with the environment;
$\widetilde{\bm{R}}_t \in \mathbb{R}^{3 \times 3}$ is a rotation matrix corresponding to the orientational component of $\widetilde{\bm{X}}_t$, 
which is used to transform the direction of force and torque to be expressed in the robot's base frame.

By integrating all the derived components above, we can give an estimated forward dynamics of the robot gripper under impedance control in a function $\Pi: \mathbb{R}^6 \times \mathbb{R}^6 \times \mathbb{R}^6 \mapsto \mathbb{R}^6 \times \mathbb{R}^6$, as expressed in Eq.~\eqref{eq:forward_dynamics}.
Given the gripper's pose $\bm{x}_t$ and velocity $\dot{\bm{x}}_t$ and the impedance control $\bm{\xi}_t$ at time $t$ (as defined in Sec.~\ref{sec:term}), the function $\Pi$ can infer the gripper's pose $\bm{x}_{t+1}$ and velocity $\dot{\bm{x}}_{t+1}$ at the next time step:
\begin{equation}
\label{eq:forward_dynamics}
    \bm{x}_{t+1}, \dot{\bm{x}}_{t+1} = \Pi \left(\bm{x}_t, \dot{\bm{x}}_t, \bm{\xi}_t \right)
\end{equation}
which is calculated by taking integrals over the time $\tau \in [t, t+1]$ as follows:
\vspace{-5pt}
\begin{subequations}
\label{eq:forward_dynamics_detail}
\begin{align}
    \bm{x}_{\tau} &= \bm{x}_t + \int_{t}^{\tau} \dot{\bm{x}}_{s} \,d s, \quad
    \dot{\bm{x}}_{\tau} = \dot{\bm{x}}_t + \int_{t}^{\tau} \ddot{\bm{x}}_{s} \,d s\\
    \ddot{\bm{x}}_{\tau} &= \bm{\Lambda}^{-1} \left( \bm{f}_{\tau} - \bm{f}_{ext} \left(\widetilde{\bm{X}}_t \right) \right)\\
    \bm{f}_{\tau} &= \bm{K} \left( \bm{\xi}_t - \bm{x}_{\tau} \right) - \bm{D} \dot{\bm{x}}_{\tau} 
\end{align}
\end{subequations}
In practice, we evaluate the above integrals numerically at small time intervals of $2$ milliseconds;
the damping parameters are set to critical damping $\bm{D} = 2\sqrt{\bm{K}}$.

Finally, by incorporating the approximated dynamics $\Pi$ as a heuristic constraint in Eq.~\eqref{eq:opt3_d}, the sequence of impedance controls can be generated by solving a constrained optimization program as expressed in Eq.~\eqref{eq:opt3}.
As visualized in Fig.~\ref{fig:impd_path}, the program aims to have the controlled robot gripper follow the constraint-exploiting geometric path $\left\{\widetilde{\bm{X}}_1, \cdots, \widetilde{\bm{X}}_T \right\}$ planned in Sec.~\ref{sec:geom_path}.
As such, the objective is to minimize $SE(3)$ distance between the planned geometric path and the predicted path from impedance-controlled dynamics.
\vspace{-5pt}
\begin{subequations}
\label{eq:opt3}
\begin{align}
    \underset{\bm{\xi}_0, \cdots, \bm{\xi}_{T-1}}{\textrm{minimize}} \quad & \sum_{t=1}^T d\left(\widetilde{\bm{X}}_t, \bm{X}_t \right) \label{eq:opt3_a}\\
	\textrm{subject to}
    \quad & \lVert \dot{\bm{x}}_0 \rVert^2 = 0, \quad \lVert \dot{\bm{x}}_T \rVert^2 = 0 \label{eq:opt3_b}\\
    \quad & \lVert \dot{\bm{x}}_t \rVert^2 \leq \delta_v, \quad \forall t = 1, \cdots, T - 1 \label{eq:opt3_c}\\
    \quad & \bm{x}_{t+1}, \dot{\bm{x}}_{t+1} = \Pi \left(\bm{x}_t, \dot{\bm{x}}_t, \bm{\xi}_t \right), \forall t = 0, \cdots, T - 1 \label{eq:opt3_d}
\end{align}
\end{subequations}
where the constraint in Eq.~\eqref{eq:opt3_b} is a boundary condition to ensure the robot gripper starts and stops at zero velocity;
Eq.~\eqref{eq:opt3_c} limits the gripper's speed by a threshold $\delta_v$ to ensure the safety of its motions.
The solution of the optimization program $\left\{\bm{\xi}_0, \cdots, \bm{\xi}_{T-1} \right\}$, as a sequence of reference equilibrium poses in $SE(3)$ for the gripper's task frame, will be commanded one by one to a Cartesian impedance controller to compliantly move the robot for execution.

\section{Experiments}
\label{sec:experiments}

\begin{figure}[t]
    \centering
    \includegraphics[width=0.95\linewidth]{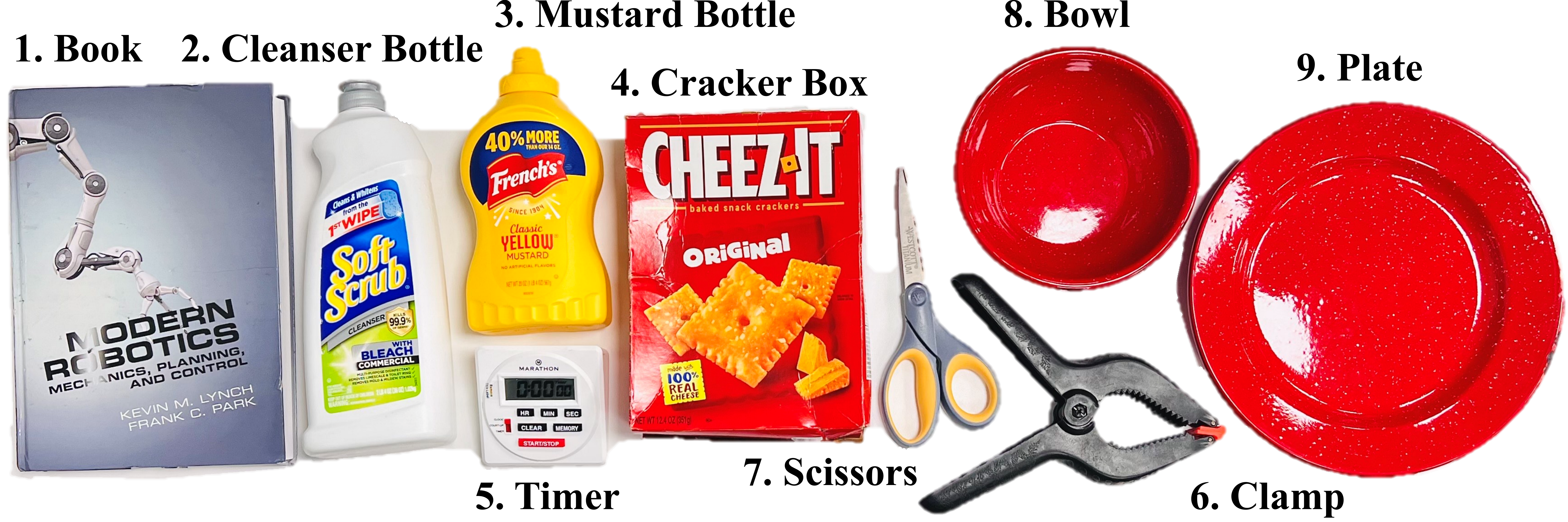}
    \vspace{-3pt}
    \caption{The objects selected for evaluation:
    1. Book~\cite{lynch2017modern};
    2. Cleanser Bottle (YCB \#20);
    3. Mustard Bottle (YCB \#9);
    4. Cracker Box (YCB \#1);
    5. Timer (YCB \#71);
    6. Clamp (YCB \#46);
    7. Scissors (YCB \#35);
    8. Bowl (YCB \#25);
    9. Plate (YCB \#24).~\cite{calli2017yale}
    }
    \label{fig:objs}
    \vspace{-10pt}
\end{figure}

We conducted real-world experiments on Franka Emika robots with both single-arm and dual-arm setups,
to evaluate our framework.
As shown in Fig.~\ref{fig:objs},
we selected $9$ objects for evaluation,
which greatly differ in geometric and physical properties such as size, weight, stiffness, etc.
We used a model-based 6D pose estimator FoundationPose~\cite{wen2024foundationpose} with CNOS~\cite{nguyen2023cnos} segmenting the objects in the first frame.
The visual models ran a single RTX 3060 GPU (12 GB).
Our collision-inclusive planner was implemented in Python with a single thread on a 3.4 GHz AMD Ryzen 9 5950X CPU;
the optimization programs (i.e., in Sec.~\ref{sec:geom_refine} and~\ref{sec:impd_ctrl}) were solved using the CasADi framework~\cite{andersson2019casadi} with IPOPT solver (an interior point method).
The length of path or trajectory was set to $T = 20$.
The SDF resolution was $\Delta=1cm$, with $\varepsilon = 1cm$ in Eq.~\eqref{eq:g_v}.
Alg.~\ref{alg:a_star} started with $\delta_g=0mm$ and incremented it by $\,d\delta_g = 1mm$ after each iteration.

\vspace{-5pt}
\subsection{Grasping Single Object on a Support Surface}
\label{sec:exp_single_obj}

\begin{figure*}[t]
    \centering
    \begin{minipage}[b]{\linewidth}
        \centering
        \includegraphics[width=0.73\linewidth]{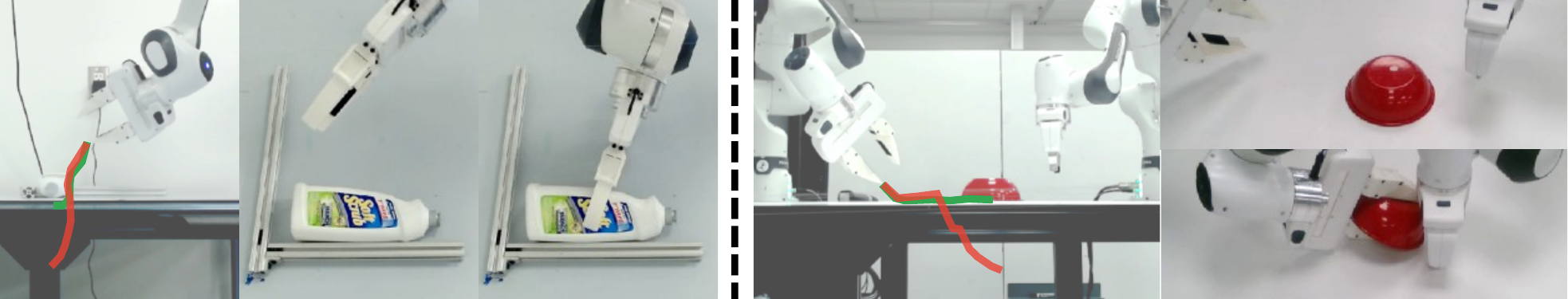}
    \end{minipage}

    \vspace{2pt}

    \begin{minipage}[b]{\linewidth}
        \centering
        \footnotesize
        \begin{tabular}{c|| c | c | c c c | c || c | c | c c c | c}
            \hline
            \multirow{3}{*}{Object} & \multicolumn{6}{c}{Single-Arm} & \multicolumn{6}{||c}{Dual-Arm} \\
            \cline{2-13}
            & \multirow{2}{*}{Success} & \multirow{2}{*}{Reps ($\pm$ std)} & \multicolumn{4}{c}{Solve Time (seconds)} & \multicolumn{1}{||c|}{\multirow{2}{*}{Success}} & \multirow{2}{*}{Reps ($\pm$ std)} & \multicolumn{4}{c}{Solve Time (seconds)}\\
            \cline{4-7} \cline{10-13}
            & & & $A^*$ & Refine & Impd & \multicolumn{1}{c}{Total ($\pm$ std)} &\multicolumn{1}{||c|}{} & & $A^*$ & Refine & Impd & Total ($\pm$ std)\\
            \hline
            Book & 10 / 10 & 1.4 $\pm$ 0.8 & 0.23 & 1.21 & 2.44 & \multicolumn{1}{c}{3.87 $\pm$ 0.68} & \multicolumn{1}{||c|}{10 / 10} & 1.1 $\pm$ 0.3 & 0.20 & 1.15 & 2.27 & 3.62 $\pm$ 2.35\\
            Cleanser & 10 / 10 & 1.1 $\pm$ 0.3 & 0.24 & 0.37 & 2.35 & \multicolumn{1}{c}{2.96 $\pm$ 0.74} & \multicolumn{1}{||c|}{10 / 10} & 1.2 $\pm$ 0.6 & 0.98 & 0.35 & 1.96 & 3.29 $\pm$ 2.72\\
            Mustard & 10 / 10 & 1.0 $\pm$ 0.0 & 0.16 & 0.46 & 2.54 & \multicolumn{1}{c}{3.16 $\pm$ 0.86} & \multicolumn{1}{||c|}{10 / 10} & 1.3 $\pm$ 0.9 & 6.87 & 0.31 & 1.93 & 9.11 $\pm$ 3.90 \\
            Cracker & 10 / 10 & 2.7 $\pm$ 1.6 & 1.87 & 1.35 & 2.16 & \multicolumn{1}{c}{5.39 $\pm$ 4.14} & \multicolumn{1}{||c|}{10 / 10} & 2.3 $\pm$ 2.0 & 3.37 & 0.16 & 2.11 & 5.64 $\pm$ 3.55 \\
            Timer & 10 / 10 & 1.1 $\pm$ 0.3 & 0.16 & 0.33 & 1.88 & \multicolumn{1}{c}{2.37 $\pm$ 0.76} & \multicolumn{1}{||c|}{10 / 10} & 1.4 $\pm$ 0.9 & 6.29 & 0.83 & 1.82 & 8.95 $\pm$ 4.52 \\
            Clamp & 10 / 10 & 1.1 $\pm$ 0.3 & 0.16 & 0.36 & 2.23 & \multicolumn{1}{c}{2.75 $\pm$ 0.61} & \multicolumn{1}{||c|}{10 / 10} & 1.2 $\pm$ 0.4 & 0.20 & 0.39 & 2.17 & 2.76 $\pm$ 0.60\\
            Scissors & 10 / 10 & 1.4 $\pm$ 0.7 & 0.13 & 2.23 & 1.99 & \multicolumn{1}{c}{4.35 $\pm$ 3.37} & \multicolumn{1}{||c|}{10 / 10} & 1.5 $\pm$ 0.8 & 0.35 & 1.21 & 1.80 & 3.36 $\pm$ 2.82\\
            Bowl & 10 / 10 & 1.6 $\pm$ 0.7 & 0.05 & 0.49 & 1.91 & \multicolumn{1}{c}{2.45 $\pm$ 0.71} & \multicolumn{1}{||c|}{10 / 10} & 3.3 $\pm$ 1.6 & 0.20 & 0.60 & 1.96 & 2.76 $\pm$ 1.46\\
            Plate & 10 / 10 & 1.0 $\pm$ 0.0 & 0.05 & 0.32 & 2.19 & \multicolumn{1}{c}{2.57 $\pm$ 0.69} & \multicolumn{1}{||c|}{10 / 10} & 1.1 $\pm$ 0.3 & 0.05 & 0.48 & 2.34 & 2.86 $\pm$ 1.26\\
            \hline
            Average & 10 / 10 & 1.4 $\pm$ 0.9 & 0.52 & 0.90 & 2.17 & \multicolumn{1}{c}{3.60 $\pm$ 2.63} & \multicolumn{1}{||c|}{10 / 10} & 1.6 $\pm$ 1.2 & 1.97 & 0.59 & 2.02 & 4.58 $\pm$ 3.66\\
            \hline
        \end{tabular}
    \end{minipage}

     \caption{
    Grasping a single object on a flat table surface by our framework, with both single-arm (left) and dual-arm (right) setups.
    The green lines show the geometric paths planned by the algorithms in Sec.~\ref{sec:geom_path}, and the red trajectories are generated impedance controls.
    The table shows evaluation results on $9$ selected objects, 
    where ``Reps'' is the average number of repetitions before success;
    the solve times of ``$A^*$'', ``Refine'', and ``Impd'' correspond to the algorithmic steps described in Sec.~\ref{sec:geom_astar}, ~\ref{sec:geom_refine}, and ~\ref{sec:impd_ctrl} respectively.
    }
    \label{fig:exp_single_obj}
    \vspace{-14pt}
\end{figure*}

With both single-arm and dual-arm setups, we evaluated our proposed framework in grasping a single object on a flat table, as exemplified in Fig.~\ref{fig:exp_single_obj}.
This task requires the robot to successfully insert one of its fingers into the narrow gap between the object and the table plane, where collisions are inevitable.
To robustly find the gap and place the finger there, 
our framework automatically exploited environmental constraints by contacting the environmental objects, as showcased in Fig.~\ref{fig:exp_single_obj}.
Our framework assumes that the environment configuration cannot be greatly changed by the robot.
To this end,
for the single-arm setup, we constructed an aluminum frame and fixed it to the table to constrain the motion range of objects;
for the dual-arm setup, we commanded one robot arm to block the object from the opposite side of the grasping direction.
The dual-arm experiments show promising feasibility of applying our framework in more realistic setups where desired constraints are not naturally available and must be created by the robot itself.
We conducted $10$ trials per object,
with a maximum of $K_{\max}=10$ repetitions per trial.
A trial was considered a failure if the object was not grasped within $10$ repetitions.

We reported the experiment results in the table of Fig.~\ref{fig:exp_single_obj}.
In all experiment trials, our framework succeeded with a small number of needed repetitions ($1.4$ for the single-arm and $1.6$ for the dual-arm setups on average), showing the high effectiveness of our framework in such task setups.
Grasping the cracker box and the bowl generally required more repetitions than other objects.
This is primarily because their thicknesses are close to the gripper’s maximum opening width, requiring more precise alignment of the gripper's orientation for successful insertion.
Furthermore, the cracker box was empty and prone to deformation, while the bowl had a narrow rim not captured in the mesh model, 
introducing unmodeled contact interactions that increased manipulation difficulty.
Although the task was not always accomplished for the first time, 
the task repetitions, 
as desirably contained in a conceptual manipulation funnel,
always produced stable outcomes and would eventually reach the grasp.
In terms of solve times, the replanning of each repetition required less than $5$ seconds.
The $A^*$-based positional path-finding (corresponding to Sec.~\ref{sec:geom_astar}) usually ran fast (i.e., less than $0.5$ seconds);
however, due to perception errors, the object may be estimated to penetrate the table with a large negative clearance between them.
Such perceptual inaccuracy could require our algorithm to run more iterations by increasing $\delta_g$ and to more exhaustively search over the entire workspace to find a feasible path, making the runtime of the iterative $A^*$ sometimes longer (e.g., more than $6$ seconds).

\begin{figure}[t]
\vspace{-5pt}
    \centering
    \begin{minipage}[b]{\linewidth}
        \centering
        \begin{tikzpicture}
            \node[anchor=south west,inner sep=0] at (0,0){\includegraphics[width=\linewidth]{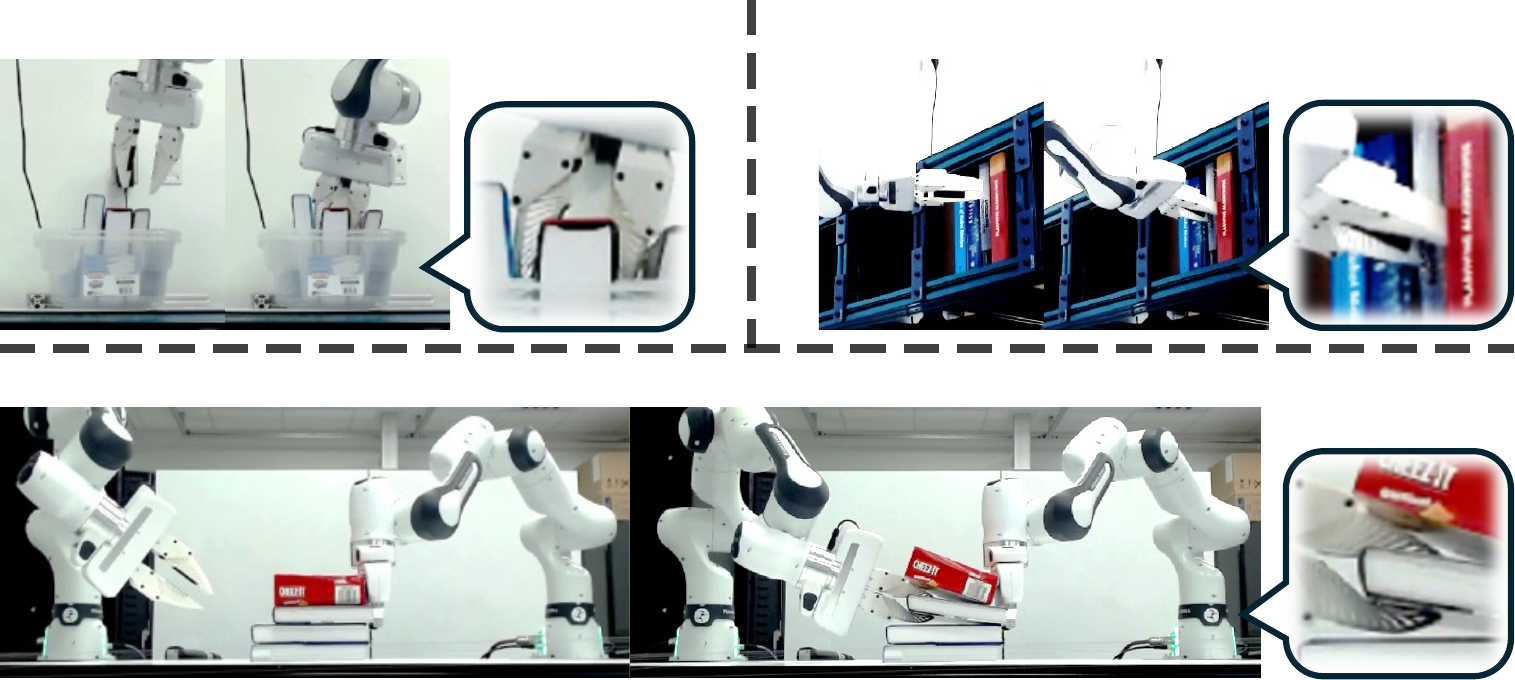}};
            \node[font=\small, anchor=west, align=left] at (0.75, 3.8) {Scene 1};
            \node[font=\footnotesize, anchor=west, align=left] at (2.58, 3.55) {Final Grasp};
            \node[font=\small, anchor=west, align=left] at (5.55, 3.8) {Scene 2};
            \node[font=\footnotesize, anchor=west, align=left] at (7.37, 3.55) {Final Grasp};
            \node[font=\small, anchor=west, align=left] at (3.1, 1.75) {Scene 3};
            \node[font=\footnotesize, anchor=west, align=left] at (7.37, 1.53) {Final Grasp};
        \end{tikzpicture}
    \end{minipage}

    \vspace{2pt}
    
    \begin{minipage}[b]{\linewidth}
    \centering
    \footnotesize
    \begin{tabular}{c|| c | c | c c c | c }
        \hline
        \multirow{2}{*}{Scene} & \multirow{2}{*}{Success} & \multirow{2}{*}{Reps ($\pm$  std)} & \multicolumn{4}{c}{Solve Time (seconds)}\\
        \cline{4 - 7}
        & & & $A^*$ & Refine & Impd & Total ($\pm$ std) \\
        \hline
        1 & 10 / 10 & 3.3 $\pm$ 1.7 & 0.14 & 2.32 & 2.10 & 4.55 $\pm$ 4.08\\
        2 & 10 / 10 & 1.1 $\pm$ 0.3 & 0.10 & 5.65 & 1.48 & 7.23 $\pm$ 6.35\\
        3 & 10 / 10 & 2.7 $\pm$ 1.6 & 1.04 & 0.64 & 2.01 & 3.69 $\pm$ 3.78\\
        \hline
    \end{tabular}
    \end{minipage}
    
    \caption{
    Grasping a target book in three scenes of tightly packed configurations.
    The table shows evaluation results on each scene.
    }
    \label{fig:exp_multi_objs}
    \vspace{-15pt}
\end{figure}

\subsection{Grasping from Tightly Stacked Multiple Objects}
For this evaluation, we further challenged our framework with more complicated environment configurations, 
where multiple objects were tightly packed in a limited space.
Unlike the single-finger insertion in Sec.~\ref{sec:exp_single_obj},
this setup required simultaneous insertion of both fingers into separate gaps,
introducing more complex collisions with the environment.
We constructed three test scenes of grasping a target book with other books (or objects) snugly next to it to occlude the desired grasp configuration, as shown in Fig.~\ref{fig:exp_multi_objs}:
1. From the tray with a single arm;
2. From the shelf with a single arm;
3. Stacked flat on the table with dual arms (the other arm blocking objects from the other side).
For each scenario, we ran $10$ trials with a maximum number of repetitions $K_{\max} = 10$.
To enable the insertion of both fingers, before execution of each trial, the robot was asked to configure its finger so that the width between fingers was similar to the thickness of the target book.

The experiment results are reported in the table of Fig.~\ref{fig:exp_multi_objs}.
As can be seen from the results, our framework still effectively succeeded in all trials of all scenarios.
Since the task setup became more complicated in terms of the environment geometries and required interactions to be considered, the average number of repetitions slightly increased.
The refinement step of geometric path planning took longer to run since the nonlinear optimizer needed more iterations to find the optimal solution for collision minimization when both fingers had to penetrate the environment.
However, the total runtime of replanning for each repetition was about $4$ to $7$ seconds in different scenes, which is still practical in reality.

\vspace{-5pt}
\subsection{Comparative Evaluation against Selected Baseline}

We selected a contact-rich baseline~\cite{wirnshofer2018robust} as it aligns more closely with our task objective, i.e., guiding the robot gripper towards a desired configuration through contacts with approximately static environment (and target object).
In contrast, 
other methods (e.g., ~\cite{cheng2023enhancing, chen2023synthesizing, zhang2023reinforcement}) focus on manipulating the target object by actively reconfiguring its pose,
and therefore may be unsuitable in highly confined spaces where object motion is restricted or undesired.

\begin{figure}[t]
    \centering
    \begin{minipage}[b]{\linewidth}
        \centering
        \begin{tikzpicture}
            \node[anchor=south west,inner sep=0] at (0,0){\includegraphics[width=0.95\linewidth]{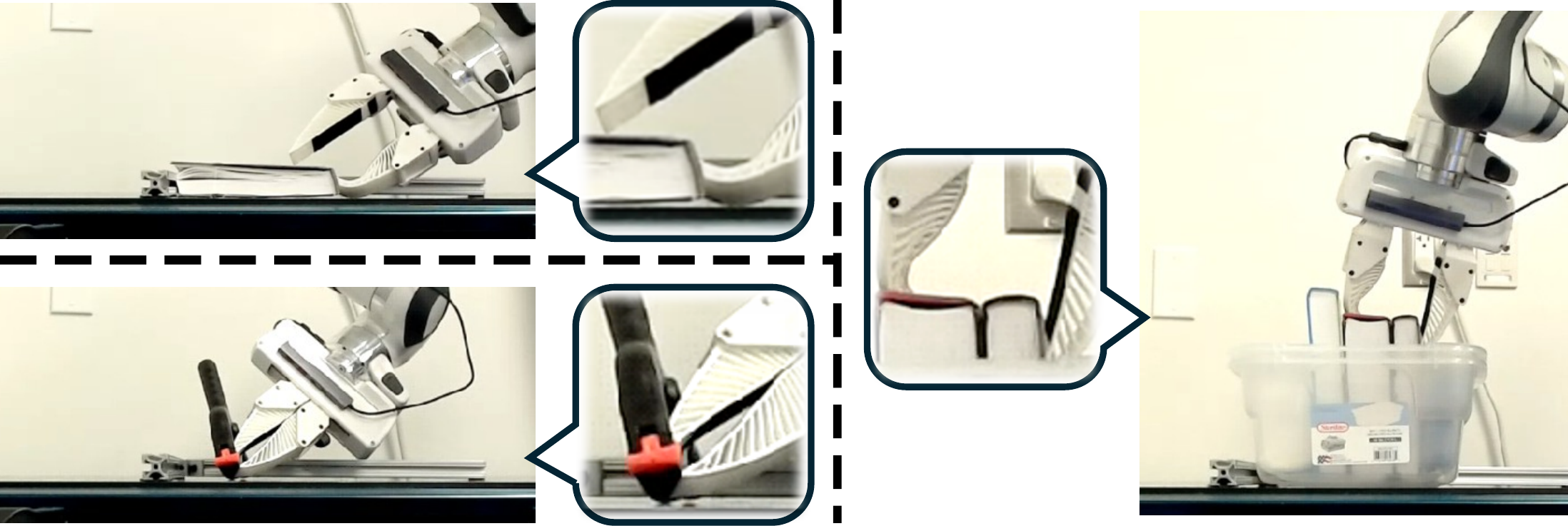}};
            \node[font=\small, anchor=west, align=left] at (0.0, 2.6) {Case 1};
            \node[font=\small, anchor=west, align=left] at (0.0, 1.1) {Case 2};
            \node[font=\small, anchor=west, align=left] at (4.51, 2.6) {Case 3};
        \end{tikzpicture}
    \end{minipage}

    \vspace{2pt}
    
    \begin{minipage}[b]{\linewidth}
    \centering
    \footnotesize
    \begin{tabular}{c  c || c  c || c c }
        \hline
        \multicolumn{2}{c ||}{Case 1} & \multicolumn{2}{c ||}{Case 2} &\multicolumn{2}{c}{Case 3} \\
        Success & Time (sec) & Success & Time (sec) & Success & Time (sec) \\
        \hline
        0 / 5 & 27.99 & 2 / 5 & 29.61 & 0 / 5 & 32.95 \\
        \hline
    \end{tabular}
    \end{minipage}    
    
    \caption{
    Failure executions by the baseline method.
    }
    \label{fig:baseline}
    \vspace{-15pt}
\end{figure}

We adapted the baseline implementation to our task setup using the same approximated dynamics model as in Eq.~\eqref{eq:forward_dynamics_detail}.
We found the baseline more sensitive to modeling inaccuracies, making it less reliable when integrated with the highly approximated dynamics. 
To avoid potential damage to objects or the other robot arm, 
we limited evaluation to three simple yet representative test cases, 
with $5$ trials each.
Each trial was given a $120$-second time budget, 
during which the baseline could replan and search for improved solutions, 
allowing multiple repetitions if time remains.
As showcased in Fig.~\ref{fig:baseline}, the baseline was more prone to execution failures,
such as unsuccessful finger insertion when grasping the book (Case 1 and 3), or incorrect grasps for the clamp (Case 2).
Additionally,
the baseline's random sampling-based motion generation required searching in a larger problem space and was inefficient,
typically requiring over $20$ seconds to find a solution, as reported in the table in Fig.~\ref{fig:baseline}.

\vspace{-5pt}
\section{Conclusion}
\label{sec:conclusion}

In this work, we proposed a collision-inclusive manipulation planning framework for collision-rich tasks using compliant robot motions, where the desired task configuration (e.g., a stable grasp for the target object) is initially occluded.
With real-world experiments on various test scenarios, 
we show that our framework can effectively address practical occluded grasping problems under uncertainties by exploiting environmental constraints and task repetitions.

\bibliographystyle{IEEEtran}
\bibliography{refs}

\begin{thebibliography}{10}
\providecommand{\url}[1]{#1}
\csname url@samestyle\endcsname
\providecommand{\newblock}{\relax}
\providecommand{\bibinfo}[2]{#2}
\providecommand{\BIBentrySTDinterwordspacing}{\spaceskip=0pt\relax}
\providecommand{\BIBentryALTinterwordstretchfactor}{4}
\providecommand{\BIBentryALTinterwordspacing}{\spaceskip=\fontdimen2\font plus
\BIBentryALTinterwordstretchfactor\fontdimen3\font minus \fontdimen4\font\relax}
\providecommand{\BIBforeignlanguage}[2]{{%
\expandafter\ifx\csname l@#1\endcsname\relax
\typeout{** WARNING: IEEEtran.bst: No hyphenation pattern has been}%
\typeout{** loaded for the language `#1'. Using the pattern for}%
\typeout{** the default language instead.}%
\else
\language=\csname l@#1\endcsname
\fi
#2}}
\providecommand{\BIBdecl}{\relax}
\BIBdecl

\bibitem{wang2019manipulation}
L.~Wang, Y.~Xiang, and D.~Fox, ``Manipulation trajectory optimization with online grasp synthesis and selection,'' in \emph{Proc. Robot.: Sci. Syst.}, 2020.

\bibitem{xiang2024grasping}
Y.~Xiang, S.~H. Allu, R.~Peddi, T.~Summers, and V.~Gogate, ``Grasping trajectory optimization with point clouds,'' in \emph{Proc. IEEE/RSJ Int. Conf. Intell. Robots Syst.}, 2024, pp. 9885--9892.

\bibitem{suomalainen2022survey}
M.~Suomalainen, Y.~Karayiannidis, and V.~Kyrki, ``A survey of robot manipulation in contact,'' \emph{Robot. Auton. Syst.}, vol. 156, p. 104224, 2022.

\bibitem{eppner2015exploitation}
C.~Eppner, R.~Deimel, J.~Alvarez-Ruiz, M.~Maertens, and O.~Brock, ``Exploitation of environmental constraints in human and robotic grasping,'' \emph{Int. J. Robot. Res.}, vol.~34, no.~7, pp. 1021--1038, 2015.

\bibitem{odhner2014compliant}
L.~U. Odhner, L.~P. Jentoft, M.~R. Claffee, N.~Corson, Y.~Tenzer, R.~R. Ma, M.~Buehler, R.~Kohout, R.~D. Howe, and A.~M. Dollar, ``A compliant, underactuated hand for robust manipulation,'' \emph{Int. J. Robot. Res.}, vol.~33, no.~5, pp. 736--752, 2014.

\bibitem{deimel2016novel}
R.~Deimel and O.~Brock, ``A novel type of compliant and underactuated robotic hand for dexterous grasping,'' \emph{Int. J. Robot. Res.}, vol.~35, no. 1-3, pp. 161--185, 2016.

\bibitem{balatti2018self}
P.~Balatti, D.~Kanoulas, G.~F. Rigano, L.~Muratore, N.~G. Tsagarakis, and A.~Ajoudani, ``A self-tuning impedance controller for autonomous robotic manipulation,'' in \emph{Proc. IEEE/RSJ Int. Conf. Intell. Robots Syst.}, 2018, pp. 5885--5891.

\bibitem{eppner2015planning}
C.~Eppner and O.~Brock, ``Planning grasp strategies that exploit environmental constraints,'' in \emph{Proc. IEEE Int. Conf. Robot. Automat.}, 2015, pp. 4947--4952.

\bibitem{mason1985mechanics}
M.~Mason, ``The mechanics of manipulation,'' in \emph{Proc. IEEE Int. Conf. Robot. Automat.}, vol.~2, 1985, pp. 544--548.

\bibitem{amanhoud2020force}
W.~Amanhoud, M.~Khoramshahi, M.~Bonnesoeur, and A.~Billard, ``Force adaptation in contact tasks with dynamical systems,'' in \emph{Proc. IEEE Int. Conf. Robot. Automat.}, 2020, pp. 6841--6847.

\bibitem{guan2018efficient}
C.~Guan, W.~Vega-Brown, and N.~Roy, ``Efficient planning for near-optimal compliant manipulation leveraging environmental contact,'' in \emph{Proc. IEEE Int. Conf. Robot. Automat.}, 2018, pp. 215--222.

\bibitem{nottensteiner2020robust}
K.~Nottensteiner, F.~Stulp, and A.~Albu-Sch{\"a}ffer, ``Robust, locally guided peg-in-hole using impedance-controlled robots,'' in \emph{Proc. IEEE Int. Conf. Robot. Automat.}, 2020, pp. 5771--5777.

\bibitem{almeida2016folding}
D.~Almeida and Y.~Karayiannidis, ``Folding assembly by means of dual-arm robotic manipulation,'' in \emph{Proc. IEEE Int. Conf. Robot. Automat.}, 2016, pp. 3987--3993.

\bibitem{dogar2011framework}
M.~R. Dogar and S.~S. Srinivasa, ``A framework for push-grasping in clutter.'' in \emph{Proc. Robot.: Sci. Syst.}, vol.~2, 2011.

\bibitem{lee2019efficient}
J.~Lee, Y.~Cho, C.~Nam, J.~Park, and C.~Kim, ``Efficient obstacle rearrangement for object manipulation tasks in cluttered environments,'' in \emph{Proc. IEEE Int. Conf. Robot. Automat.}, 2019, pp. 183--189.

\bibitem{king2013pregrasp}
J.~E. King, M.~Klingensmith, C.~M. Dellin, M.~R. Dogar, P.~Velagapudi, N.~S. Pollard, and S.~S. Srinivasa, ``Pregrasp manipulation as trajectory optimization.'' in \emph{Proc. Robot.: Sci. Syst.}, 2013.

\bibitem{hang2019pre}
K.~Hang, A.~S. Morgan, and A.~M. Dollar, ``Pre-grasp sliding manipulation of thin objects using soft, compliant, or underactuated hands,'' \emph{IEEE Robot. Automat. Lett.}, vol.~4, no.~2, pp. 662--669, 2019.

\bibitem{liang2021learning}
H.~Liang, X.~Lou, Y.~Yang, and C.~Choi, ``Learning visual affordances with target-orientated deep q-network to grasp objects by harnessing environmental fixtures,'' in \emph{Proc. IEEE Int. Conf. Robot. Automat.}, 2021, pp. 2562--2568.

\bibitem{sun2020learning}
Z.~Sun, K.~Yuan, W.~Hu, C.~Yang, and Z.~Li, ``Learning pregrasp manipulation of objects from ungraspable poses,'' in \emph{Proc. IEEE Int. Conf. Robot. Automat.}, 2020, pp. 9917--9923.

\bibitem{zhou2023learning}
W.~Zhou and D.~Held, ``Learning to grasp the ungraspable with emergent extrinsic dexterity,'' in \emph{Conf. Robot. Learn.}, 2023, pp. 150--160.

\bibitem{yang2024learning}
S.-M. Yang, M.~Magnusson, J.~A. Stork, and T.~Stoyanov, ``Learning extrinsic dexterity with parameterized manipulation primitives,'' in \emph{Proc. IEEE Int. Conf. Robot. Automat.}, 2024, pp. 5404--5410.

\bibitem{cheng2023enhancing}
X.~Cheng, S.~Patil, Z.~Temel, O.~Kroemer, and M.~T. Mason, ``Enhancing dexterity in robotic manipulation via hierarchical contact exploration,'' \emph{IEEE Robot. Automat. Lett.}, vol.~9, no.~1, pp. 390--397, 2023.

\bibitem{chen2023synthesizing}
S.~Chen, A.~Wu, and C.~K. Liu, ``Synthesizing dexterous nonprehensile pregrasp for ungraspable objects,'' in \emph{ACM SIGGRAPH 2023 Conference Proceedings}, 2023, pp. 1--10.

\bibitem{zhang2023reinforcement}
H.~Zhang, H.~Liang, L.~Cong, J.~Lyu, L.~Zeng, P.~Feng, and J.~Zhang, ``Reinforcement learning based pushing and grasping objects from ungraspable poses,'' in \emph{Proc. IEEE Int. Conf. Robot. Automat.}, 2023, pp. 3860--3866.

\bibitem{sieverling2017interleaving}
A.~Sieverling, C.~Eppner, F.~Wolff, and O.~Brock, ``Interleaving motion in contact and in free space for planning under uncertainty,'' in \emph{Proc. IEEE/RSJ Int. Conf. Intell. Robots Syst.}, 2017, pp. 4011--4073.

\bibitem{pall2021analysis}
E.~P{\'a}ll and O.~Brock, ``Analysis of open-loop grasping from piles,'' in \emph{Proc. IEEE Int. Conf. Robot. Automat.}, 2021, pp. 2591--2597.

\bibitem{dafle2014extrinsic}
N.~C. Dafle, A.~Rodriguez, R.~Paolini, B.~Tang, S.~S. Srinivasa, M.~Erdmann, M.~T. Mason, I.~Lundberg, H.~Staab, and T.~Fuhlbrigge, ``Extrinsic dexterity: In-hand manipulation with external forces,'' in \emph{Proc. IEEE Int. Conf. Robot. Automat.}, 2014, pp. 1578--1585.

\bibitem{subramani2018inferring}
G.~Subramani, M.~Zinn, and M.~Gleicher, ``Inferring geometric constraints in human demonstrations,'' in \emph{Conf. Robot. Learn.}, 2018, pp. 223--236.

\bibitem{shao2020learning}
L.~Shao, T.~Migimatsu, and J.~Bohg, ``Learning to scaffold the development of robotic manipulation skills,'' in \emph{Proc. IEEE Int. Conf. Robot. Automat.}, 2020, pp. 5671--5677.

\bibitem{li2022learning}
X.~Li and O.~Brock, ``Learning from demonstration based on environmental constraints,'' \emph{IEEE Robot. Automat. Lett.}, vol.~7, no.~4, pp. 10\,938--10\,945, 2022.

\bibitem{ratliff2009chomp}
N.~Ratliff, M.~Zucker, J.~A. Bagnell, and S.~Srinivasa, ``Chomp: Gradient optimization techniques for efficient motion planning,'' in \emph{Proc. IEEE Int. Conf. Robot. Automat.}, 2009, pp. 489--494.

\bibitem{wen2024foundationpose}
B.~Wen, W.~Yang, J.~Kautz, and S.~Birchfield, ``Foundationpose: Unified 6d pose estimation and tracking of novel objects,'' in \emph{Proc. IEEE/CVF Conf. Comput. Vis. Pattern Recognit.}, 2024, pp. 17\,868--17\,879.

\bibitem{nguyen2023cnos}
V.~N. Nguyen, T.~Groueix, G.~Ponimatkin, V.~Lepetit, and T.~Hodan, ``Cnos: A strong baseline for cad-based novel object segmentation,'' in \emph{Proc. IEEE/CVF Int. Conf. Comput. Vis.}, 2023, pp. 2134--2140.

\bibitem{andersson2019casadi}
J.~A. Andersson, J.~Gillis, G.~Horn, J.~B. Rawlings, and M.~Diehl, ``Casadi: a software framework for nonlinear optimization and optimal control,'' \emph{Mathematical Programming Computation}, vol.~11, pp. 1--36, 2019.

\bibitem{lynch2017modern}
K.~Lynch, \emph{Modern Robotics}.\hskip 1em plus 0.5em minus 0.4em\relax Cambridge University Press, 2017.

\bibitem{calli2017yale}
B.~Calli, A.~Singh, J.~Bruce, A.~Walsman, K.~Konolige, S.~Srinivasa, P.~Abbeel, and A.~M. Dollar, ``Yale-cmu-berkeley dataset for robotic manipulation research,'' \emph{Int. J. Robot. Res.}, vol.~36, no.~3, pp. 261--268, 2017.

\bibitem{wirnshofer2018robust}
F.~Wirnshofer, P.~S. Schmitt, W.~Feiten, G.~v. Wichert, and W.~Burgard, ``Robust, compliant assembly via optimal belief space planning,'' in \emph{Proc. IEEE Int. Conf. Robot. Automat.}, 2018, pp. 5436--5443.

\end{thebibliography}

\end{document}